\documentclass{cstr}

\usepackage{amsmath,amsfonts,amssymb,latexsym,bm}
\usepackage{amsthm}
\usepackage{mathrsfs}
\usepackage[dvipdfmx]{graphicx}
\usepackage{subfig}
\usepackage{url}
\usepackage{cite}
\usepackage{color}

\usepackage{enumerate}

\usepackage{algorithm, algorithmic}
\renewcommand{\algorithmicrequire}{{\bf Input:}}
\renewcommand{\algorithmicensure}{{\bf Output:}}

\title{
        Data collaboration analysis for distributed datasets
}

\author[1,*]{Akira Imakura}
\author[1]{Tetsuya Sakurai}

\affil[1]{University of Tsukuba}

\email{imakura@cs.tsukuba.ac.jp}

\begin{document}
\maketitle
\thispagestyle{titlepage}

\begin{abstract}
In this paper, we propose a {\it data collaboration analysis} method for distributed datasets.
The proposed method is a centralized machine learning while training datasets and models remain distributed over some institutions.
Recently, data became large and distributed with decreasing costs of data collection.
If we can centralize these distributed datasets and analyse them as one dataset, we expect to obtain novel insight and achieve a higher prediction performance compared with individual analyses on each distributed dataset.
However, it is generally difficult to centralize the original datasets due to their huge data size or regarding a privacy-preserving problem.
To avoid these difficulties, we propose a data collaboration analysis method for distributed datasets without sharing the original datasets.
The proposed method centralizes only {\it intermediate representation} constructed individually instead of the original dataset.
\end{abstract}

\section{Introduction}
Recently, data became large and distributed with decreasing costs of data collection.
If we can centralize these distributed datasets and analyse them as one dataset, we call this as {\it centralized analysis}, we expect to obtain novel insight and achieve higher prediction performance compared with the {\it individual analysis} on each distributed dataset.
However, it is generally difficult to centralize the original datasets due to its huge data size or regarding a privacy-preserving problem.
\par
As an example, in the case of medical data analyses, the dataset in each institution may not be enough for the high prediction performance due to deficiency and imbalance of the data samples.
However, it is difficult to centralize the original medical data samples with other institutions due to a privacy-preserving problem.
On the other hand, it is highly possible that, if the original data is transformed to other (low-)dimensional space by some appropriate linear/nonlinear map, the mapped data called {\it intermediate representation} can be centralized relatively easily because each features of intermediate representation does not have any physical meaning.
\par
Examples to overcome the difficulties of the centralized analysis include usage of privacy-preserving computation based on cryptography \cite{jha2005privacy,kerschbaum2012privacy,cho2018secure,gilad2016cryptonets} and differential privacy \cite{abadi2016deep,ji2014differential,dwork2006differential}.
The federated learning \cite{konevcny2016federated,mcmahan2016communication}, that centralizes a model while the original datasets remains distributed has also been studied for this context.
\par
As a different approach from these existing methods, in this paper, we propose a {\it data collaboration analysis} method for distributed datasets based on centralizing only intermediate representation constructed individually.
The proposed data collaboration analysis method assumes that each institution uses a different map function for constructing intermediate representation and does not centralize the map functions to avoid risk for approximating the original data samples from intermediate representation using (approximate) inverse of the map function.
The proposed data collaboration analysis method also does not use privacy-preserving computation. 
Using some sharable data, e.g., public data and dummy data constructed randomly, as anchor, the proposed method realizes a data collaboration analysis by mapping again the individual intermediate representation to the same incorporable representation named {\it collaboration representation}.
\par
The main contributions of this paper are
\begin{itemize}
        \item We propose the data collaboration analysis method without centralizing the original dataset for distributed datasets.
        \item The proposed method is different from the existing approaches because it does not use privacy-preserving computations and does not centralize a model.
\end{itemize}
\section{Data collaboration analysis}
Here, we consider the case that there are multiple institutions and each institution has dataset individually.
We propose a data collaboration analysis method for distributed datasets without sharing the original data.
In this paper, we assume that analysis means supervised learning including classification and regression.
\par
Let $d$ be the number of institutions, $m$ be the number of features, $n_i, s_i$ be the numbers of training/test data samples the $i$th institution has and $n, s$ be the total numbers of training/test data samples, $n = \sum_{i=1}^d n_i, s = \sum_{i=1}^d s_i$.
We also let $X_i = [{\bm x}_{i1}, {\bm x}_{i2}, \dots, {\bm x}_{in_i}] \in \mathbb{R}^{m \times n_i}$, $L_i = [{\bm l}_{i1}, {\bm l}_{i2}, \dots, {\bm l}_{in_i}] \in \mathbb{R}^{l \times n_i}$ and $Y_i = [{\bm y}_{i1}, {\bm y}_{i2}, \dots, {\bm y}_{is_i}] \in \mathbb{R}^{m \times s_i}$ be training dataset, ground truth and test dataset the $i$th institution has.
\par
We do not centralize the original datasets $X_i$ and $Y_i$.
Instead, we centralize the intermediate representation constructed individually from $X_i$.
We also do not centralize the map function for the intermediate representation to reduce risk for approximating the original data.
\par
In the rest of this section, we introduce the basic concept and propose a practical algorithm.
\subsection{The basic concept}
Instead of centralizing the original dataset $X_i$, we consider centralizing the intermediate representation,
\begin{align*}
        \widetilde{X}_i = [\widetilde{\bm x}_{i1}, \widetilde{\bm x}_{i2}, \dots, \widetilde{\bm x}_{in_i}] = f_i(X_i) \in \mathbb{R}^{\ell_i \times n_i}, \quad
        \widetilde{\bm x}_{ik} = f_i({\bm x}_{ik})  \quad
        (1 \leq i \leq d, 1 \leq k \leq n_i)
\end{align*}
constructed in each institution individually, where $f_i$ is a linear/nonlinear column-wise map function.
Since each map function $f_i$ is constructed using $X_i$ individually, it depends on $i$.
The dimensionality $\ell_i$ may also vary the institution $i$.
\par
The map function includes unsupervised dimensionality reductions such as Principal component analysis (PCA) \cite{pearson1901liii,jolliffe1986principal}, locality preserving projections (LPP) \cite{he2004locality} and supervised dimensionality reductions such as Fisher discriminant analysis (FDA) \cite{fisher1936use,fukunaga2013introduction}, local FDA (LFDA) \cite{sugiyama2007dimensionality}, semi-supervised LFDA (SELF) \cite{sugiyama2010semi}, locality adaptive discriminant analysis (LADA) \cite{li2017locality}, complex moment-based supervised eigenmap (CMSE) \cite{imakura2019complex}.
We also consider a partial structure of deep neural networks.
The proposed method aims to avoid a privacy-preserving problem by achieving collaboration analysis while the original dataset $X_i$ and the map function $f_i$ remain distributed in each institution.
\par
Because $f_i$ depends on the institution $i$, even if each institution has the same data sample ${\bm x}$, its intermediate representation is different, i.e.,
\begin{equation*}
        f_i({\bm x}) \neq f_j({\bm x}) \quad
        (i \neq j).
\end{equation*}
Therefore, we can not analyse intermediate representations as one dataset even if dimensionality is the same, $\ell_i = \ell_j$.
\par
To overcome this difficulty, we consider transforming again the individual intermediate representations to the same incorporable representation by a linear map, i.e.,
\begin{align*}
        \widehat{X}_i = [\widehat{\bm x}_{i1}, \widehat{\bm x}_{i2}, \dots, \widehat{\bm x}_{in_i}] = G_i \widetilde{X}_i \in \mathbb{R}^{\ell \times n_i}, \quad
        \widehat{\bm x}_{ik} = G_i \widetilde{\bm x}_{ik} \quad (1 \leq i \leq d, 1 \leq k \leq n_i)
\end{align*}
with $G_i \in \mathbb{R}^{\ell \times \ell_i}$ such that
\begin{align*}
        G_i f_i({\bm x}) \approx G_j f_j({\bm x}) \quad
        (i \neq j).
\end{align*}
Since the obtained data $\widehat{\bm x}_{ik}$ $(1 \leq i \leq d, 1 \leq k \leq n_i)$ preserve some relationship of the original dataset, we can analyse them as one dataset
\begin{equation*}
        \widehat{X} = [\widehat{X}_1, \widehat{X}_2, \dots, \widehat{X}_{d}] \in \mathbb{R}^{\ell \times n}.
\end{equation*}
\par
Because the map function $f_i$ for the intermediate representations are not centralized, the matrix $G_i$ can not be constructed from only the centralized intermediate representation $\widetilde{X}_i$.
To construct the matrix $G_i$, here, we introduce a sharable data called {\it anchor dataset},
\begin{equation*}
        X^{\rm anc} = [{\bm x}_1^{\rm anc}, {\bm x}_2^{\rm anc}, \dots, {\bm x}_r^{\rm anc}] \in \mathbb{R}^{m \times r},
\end{equation*}
e.g., public data and dummy data constructed randomly, where $r \geq \ell_i$.
Applying each map function $f_i$ to the anchor data, we have the $i$th intermediate representation of the anchor dataset,
\begin{equation*}
        \widetilde{X}_i^{\rm anc} = [\widetilde{\bm x}_{i1}^{\rm anc}, \widetilde{\bm x}_{i2}^{\rm anc}, \dots, \widetilde{\bm x}_{ir}^{\rm anc}] = f_i(X^{\rm anc}) \in \mathbb{R}^{\ell_i \times r}.
\end{equation*}
Then, we construct the matrix $G_i$ such that the individual intermediate representation of the anchor data
\begin{equation*}
        \widehat{X}_i^{\rm anc} = [\widehat{\bm x}_{i1}^{\rm anc}, \widehat{\bm x}_{i2}^{\rm anc}, \dots, \widehat{\bm x}_{in_i}^{\rm anc}] = G_i \widetilde{X}_i^{\rm anc} \in \mathbb{R}^{\ell \times r}
\end{equation*}
satisfy
\begin{equation*}
        \widehat{\bm x}_{ik}^{\rm anc} \approx \widehat{\bm x}_{jk}^{\rm anc} \quad
        (i \neq j).
\end{equation*}
\subsection{Proposal for a practical algorithm}
One of the main parts of the proposed method is constructing the collaboration representation.
The matrix $G_i$ can be constructed via the following two steps: setting the target for the collaboration representation and constructing the matrix $G_i$.
\par
Firstly, we set the target $Z = [{\bm z}_1, {\bm z}_2, \dots, {\bm z}_r] \in \mathbb{R}^{\ell \times r}$ for the collaboration representation $\widehat{X}_i^{\rm anc}$ of the anchor data by solving
\begin{equation*}
        \min_{G_1', G_2', \dots, G_d', Z} \sum_{i=1}^d \| Z - G_i' \widetilde{X}_{i}^{\rm anc} \|_{\rm F}^2.
\end{equation*}
This minimization problem is difficult to solve directly.
Instead, we consider solving the following minimal perturbation problem, i.e.,
\begin{equation}
        \min_{E_i, G_i' (i = 1, 2, \dots, d), Z} \sum_{i=1}^d \| E_i \|_{\rm F}^2 \quad \mbox{s.t. } G_i' (\widetilde{X}_{i}^{\rm anc} + E_i) = Z.
        \label{eq:tls}
\end{equation}
\par
The minimal perturbation problem \eqref{eq:tls} with $d = 2$ is called the total least squares problem and is solved by the singular value decomposition (SVD) \cite{ito2016algorithm}.
As the same manner, we can solve \eqref{eq:tls} with $d > 2$ using SVD.
Let 
\begin{align}
        [(\widetilde{X}^{\rm anc}_1)^{\rm T}, (\widetilde{X}^{\rm anc}_2)^{\rm T}, \dots, (\widetilde{X}^{\rm anc}_d)^{\rm T}] 
        = \left[ U_1, U_2 \right] \left[ \begin{array}{cc}
                        \Sigma_1 & \\
                        & \Sigma_2 
                \end{array}
        \right] \left[ \begin{array}{cccc}
                        V_{11}^{\rm T} & V_{21}^{\rm T} & \dots & V_{d1}^{\rm T} \\
                        V_{12}^{\rm T} & V_{22}^{\rm T} & \dots & V_{d2}^{\rm T}
                \end{array}
        \right]
        \label{eq:svd}
\end{align}
be SVD of the matrix combining $\widetilde{X}_i^{\rm anc}$, where
\begin{equation*}
        U_1 \in \mathbb{R}^{r \times \ell}, \quad
        \Sigma_1 \in \mathbb{R}^{\ell \times \ell}, \quad
        V_{i1} \in \mathbb{R}^{\ell_i \times \ell},
\end{equation*}
and $\Sigma_1$ has larger part of singular values.
Then, we have
\begin{equation*}
        Z = C U_1^{\rm T},
\end{equation*}
where $C\in \mathbb{R}^{\ell \times \ell}$ is a nonsingular matrix.
\par
Next, setting $Z = U_1^{\rm T}$, we construct the matrix $G_i$ such that
\begin{equation*}
        G_i = \arg \min_{G} \| Z - G \widetilde{X}_i^{\rm anc} \|_{\rm F}^2 = U_1^{\rm T} (\widetilde{X}_i^{\rm anc})^\dagger.
%        \label{eq:lsq}
\end{equation*}
\par
The algorithm of the proposed method is summarised in Algorithm~\ref{alg:proposed}.
\begin{algorithm}[t]
\caption{A proposed method}
\label{alg:proposed}
\begin{algorithmic}[1]
        \REQUIRE $X_i \in \mathbb{R}^{m \times n_i}, L_i \in \mathbb{R}^{l \times n_i}, Y_i \in \mathbb{R}^{m \times s_i}$ $(i = 1, 2, \dots, d)$ individually.
        \ENSURE $L_{Y_i} \in \mathbb{R}^{l \times s_i}$ $(i = 1, 2, \dots, d)$ individually.
        \\ \COMMENT{Phase 0. Preparation}
        \STATE Centralize $X^{\rm anc} \in \mathbb{R}^{m \times r}$
        \\[12pt] \COMMENT{Phase 1. Individual preparation}
        \STATE Construct $\widetilde{X}_i = f_i(X_i)$ and $X_i^{\rm anc} = f_i(X^{\rm anc})$ for each $i$ individually \\
        \STATE Centralize $\widetilde{X}_i, \widetilde{X}_i^{\rm anc}, L_i$ for all $i$
        \\[12pt] \COMMENT{Phase 2. Data collaboration}
        \STATE Compute left singular vectors $U_1$ of SVD \eqref{eq:svd}
        \STATE Compute $G_i = U_1^{\rm T} (\widetilde{X}_i^{\rm anc})^\dagger$
        \STATE Compute $\widehat{X}_{i} = G_i \widetilde{X}_{i}$
        \STATE Set $\widehat{X} = [\widehat{X}_1, \widehat{X}_2, \dots, \widehat{X}_d]$ and $L = [L_1, L_2, \dots, L_d]$
        \\[12pt]
        \COMMENT{Phase 3. Analysis}
        \STATE Construct a model $h$ by some machine learning or deep learning algorithm using $\widehat{X}$ as the training date and $L$ as the ground truth, i.e., $L \approx h(\widehat{X})$.
        \STATE Predict the test data $\widehat{Y}_i$ using a model $h$ and get $L_{Y_i} = h(G_if_i({Y}_i))$.
\end{algorithmic}
\end{algorithm}
\subsection{Related works}
One possible choice to achieve a high quality analysis avoiding the difficulties of the centralized analysis is usage of privacy-preserving computation.
There are two typical types of the privacy-preserving computation techniques based on cryptography \cite{jha2005privacy,kerschbaum2012privacy,cho2018secure,gilad2016cryptonets} and differential privacy \cite{abadi2016deep,ji2014differential,dwork2006differential}.
\par
Cryptography-type privacy-preserving computations (or secure multi-party computations) can compute a function over distributed data whole keeping those data private.
Specifically, fully homomorphic encryption (FHE) \cite{gentry2009fully} can compute any given function; however, they are impractical for large datasets regarding computational costs even using the latest implementations \cite{chillotti2016faster}.
The differential privacy is another type of privacy-preserving computations that protects privacy of the original datasets by randomization.
The differential privacy-type privacy-preserving computations are more efficient than cryptography-type regarding the computational costs, however, they may be low prediction accuracy due to noise added for protecting privacy.
\par
As another choice, the federated learning that centralizing a model has also been studied \cite{konevcny2016federated,mcmahan2016communication}.
The federate learning achieves a high quality analysis avoiding the difficulties of the centralized analysis by centralizing a model function instead of using cryptography and randomization.
However, since it may have a risk for revealing the original dataset because a model is shared.
Therefore, the federated learning is used in conjunction with privacy-preserving computations \cite{yang2019gdpr}.
\par
The proposed method is different from these existing approaches because it does not use privacy-preserving computations and does not centralize a model.
\section{Conclusions}
In this paper, we proposed a {\it data collaboration analysis} method for distributed datasets based on centralizing only individual intermediate representation while the original datasets and the map functions remain distributed.
The proposed method is different from existing approaches, since it does not use the privacy-preserving computations and does not centralize the map functions.
\par
In the future, we will evaluate the recognition performance of the proposed method compared with the centralized and individual analyses for several problems.
\section{Acknowledgments}
The present study is supported in part by the Japan Science and Technology Agency (JST), ACT-I (No. JPMJPR16U6), the New Energy and Industrial Technology Development Organization (NEDO) and the Japan Society for the Promotion of Science (JSPS), Grants-in-Aid for Scientific Research (Nos.~17K12690, 18H03250).
\bibliography{mybibfile}
\bibliographystyle{elsart-num-sort}
\end{document}